\documentclass[10pt,twocolumn,letterpaper]{article}

\usepackage[pagenumbers]{cvpr} 
\usepackage{graphicx}
\usepackage{booktabs}

\usepackage{colortbl}
\definecolor{tabfirst}{HTML}{F4A6A3}
\definecolor{tabsecond}{HTML}{F8CFA6}
\newcommand{\first}[1]{\cellcolor{tabfirst}#1}
\newcommand{\second}[1]{\cellcolor{tabsecond}#1}
\definecolor{cvprblue}{rgb}{0.21,0.49,0.74}
\usepackage[pagebackref,breaklinks,colorlinks,allcolors=cvprblue]{hyperref}

\def\confName{CVPR}
\def\confYear{2027}

\title{HumanBridge: Bridging Dissimilar Human Motions with Contact Planning}

\author{Fa-Ting Hong \qquad Xiangjun Tang \qquad Peter Wonka\\
King Abdullah University of Science and Technology (KAUST)\\
{\tt\small fating.hong@kaust.edu.sa \quad xiangjun.tang@kaust.edu.sa \quad peter.wonka@kaust.edu.sa}
}

\begin{document}
\twocolumn[{
\renewcommand\twocolumn[1][]{#1}
\renewcommand{\thefootnote}{\fnsymbol{footnote}}
\maketitle
\begin{center}
    \centering
    \captionsetup{type=figure}
    \includegraphics[width=0.98\textwidth]{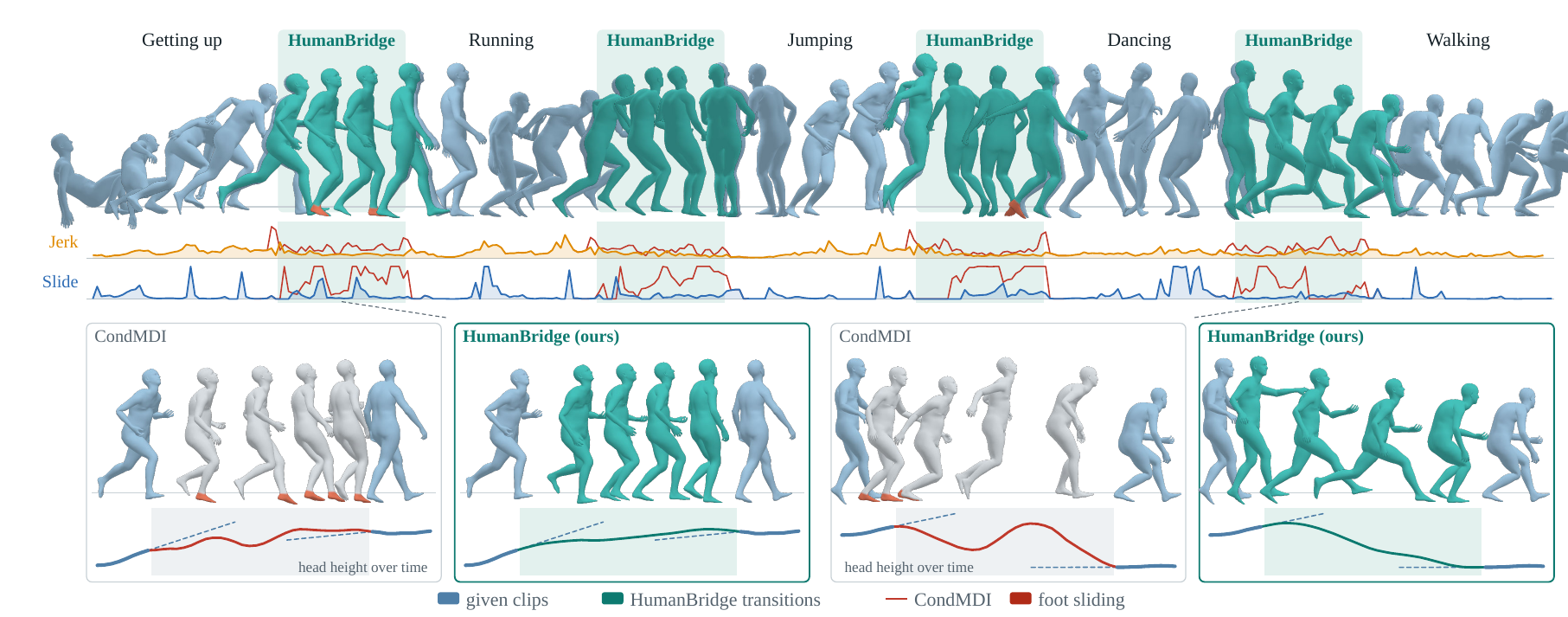}
    \vspace{-10pt}
    \captionof{figure}{HumanBridge joins five existing motion clips (blue) into one animation without changing them. Compared with CondMDI~\cite{cohan2024flexible}, our transitions join both clips without jumps and keep the feet on the floor. Below each close-up, we plot the height of the head over time, where the dashed lines continue each clip at its own velocity. Our transition leaves along the first clip and arrives along the second.
    }
    \label{fig:teaser}
\end{center}
}]

\begin{abstract}
Connecting two existing motions is a fundamental operation in character animation. Both clips must be played back unchanged, yet they usually come from different motions and disagree in pose, velocity, and gait phase, so no ground-truth transition exists. Inertialization keeps the boundaries smooth, but it only blends the offset between the clips, so the feet slide and sink into the floor. In-betweening and motion diffusion methods instead synthesize new motion, but they produce visible jumps at the boundaries. To address both problems, we introduce HumanBridge, a diffusion-based framework that bridges two existing motions with seamless transitions and high-quality foot motion. At the boundaries, two learned blending schedules guide each denoising step toward extrapolations of the prefix and suffix clips, which preserves velocity continuity. Within the transition, a contact planner predicts a foot contact schedule that continues the stepping pattern of the prefix clip into that of the suffix clip, and lightweight cross-attention adapters condition every transformer block on this schedule. Since no ground truth is available, we train the model through its complete sampling chain, with losses on boundary continuity, smoothness, foot skating and floor penetration. Experiments on LaFAN1 and HumanML3D show that HumanBridge is the only method that keeps both the boundaries continuous and the foot motion close to real motion. On LaFAN1, compared with inertialization, it reduces foot skating by 4.6 times and floor penetration from 13.47\% to 0.23\%, while keeping boundary discontinuities at 0.2\%. Ablation studies show that the blending schedules reduce both the jumps at the boundaries and foot skating, and that without the contact plan both feet stay off the ground in 86.0\% of the frames.

\end{abstract}
    
\section{Introduction}
\label{sec:intro}

Transitions between existing motion clips are everywhere in character animation. Game engines switch between motion clips whenever the player's input or the character's state changes, motion graphs and motion matching continually jump from one captured segment to another \cite{kovar2023motion}, and editors stitch captured takes into longer sequences. In all of these applications, the two motion clips are fixed, and the transition is therefore the only part of the motion that has to be generated, and its quality determines whether the character appears to move continuously or visibly pops, slides, and floats between two given high-quality motion clips.

In practice, such transitions are still produced with simple, hand-crafted techniques. Crossfade blending and inertialization \cite{bollo2018inertialization} remove the discontinuity at the switch by blending or decaying the offset between the two clips, which is fast and robust, but it only smooths the offset between motion clips away rather than synthesizing new motion. The feet slide when the given motion clips are in different patterns of their steps, and long transitions degenerate into an interpolation of the two poses. Locking the feet with inverse kinematics reduces the sliding, but the locked feet then have to be released abruptly to reach the next clip, which brings the jumps back. Learned motion in-betweening methods \cite{harvey2020robust,qin2022motion,tang2022real,oreshkin2023motion, akhoundi2025silk,tang2023rsmt} generate realistic motion between a context and a target keyframe. However, these methods are trained by masking out frames within continuous motion capture sequences, where the context and the keyframe belong to the same motion and the masked frames serve as ground truth. When bridging two existing clips, the two sides typically come from different motions and disagree in pose, velocity and foot contacts. That means the inputs fall outside the training distribution, and no ground-truth transition exists to supervise the training.

Most recently, diffusion models have shown strong capabilities for generating diverse and realistic human motion \cite{tevet2022human,rempe2026kimodo,zhao2026autoregressive}, and their flexibility makes it natural to cast the transition as a completion problem. The frames of both motion clips are given as observations, and the frames in between are generated by inpainting or by conditioning on a mask \cite{cohan2024flexible,shafir2024human,barquero2024seamless, agrawal2026interactive}. While these methods produce plausible motion inside the gap, the observed frames only enter as inputs, and nothing ensures that the generated motion leaves the prefix motion clip and enters the suffix motion clip with matching velocity. As a result, they frequently produce visible jumps at the transition boundaries. Moreover, without an explicit notion of when the feet should be on the ground, the generated transitions tend to slide or keep both feet off the ground for long stretches. Bridging two existing clips therefore calls for a transition that connects smoothly to both given motion clips at its boundaries, while preventing the feet from sliding and penetrating the ground (Figure~\ref{fig:teaser}).

In this work, we introduce HumanBridge, a diffusion-based framework that bridges two existing motion clips with seamless transitions and high-quality foot motion, while playing both motion clips back unchanged. HumanBridge fine-tunes a conditional diffusion transformer and comprises three main components. First, at the transition boundaries, two learned blending schedules softly guide the clean-motion prediction at each denoising step toward extrapolations of the prefix and suffix motion clips. The schedules depend on the position within the transition and on the noise level, so they can enforce velocity continuity at both boundaries while leaving the middle of the transition to the generative model. Second, within the transition, a contact planner predicts from both clips a foot contact schedule that continues the stepping pattern of the prefix clip into that of the suffix clip. Lightweight cross-attention adapters condition every transformer block on this schedule and on each frame's position within the transition, so that the plan directly conditions the denoiser at every layer, not only through the training objective. Third, since no ground-truth transition exists, we train the model by running the full denoising rollout. Every training sample is generated from pure noise with the same sampler used at inference, and losses on boundary continuity, smoothness, foot skating, and floor penetration are applied to the final result. This lets the blending schedules, the adapters, and the denoiser learn jointly from the quality of the transitions they produce, without any paired transition data.

We evaluate HumanBridge on a bridging protocol built from LaFAN1 \cite{harvey2020robust}, in which the prefix and suffix clips are drawn from different motions, and repeat the full protocol on HumanML3D \cite{guo2022generating} to assess generality. We compare against inertialization~\cite{bollo2018inertialization}, learned in-betweening~\cite{qin2022motion}, and recent motion diffusion and inpainting methods~\cite{shafir2024human,barquero2024seamless,cohan2024flexible,rempe2026kimodo,agrawal2026interactive,nazarenus2026actionplan}, measuring discontinuities at both boundaries, foot skating, foot contact statistics, and floor penetration. HumanBridge is the only method that keeps both boundaries continuous and the feet high-quality foot motion relative to the ground. Inertialization connects the clips smoothly but lets the feet slide and sink into the floor, locking its feet brings the jumps back, and the diffusion methods jump at the boundaries. Moreover, we validate our key design choices through extensive ablations, which reveal that seamless boundaries stem from the two-sided blending schedule and full-chain training, whereas stable foot contact emerges from the interplay of the contact planner and its adapters. 

In summary, the key contributions are as follows:
\begin{itemize}
    \item We propose HumanBridge, a diffusion-based framework that bridges two existing motion clips while playing both back unchanged, and train it through its complete sampling chain without requiring ground-truth transitions.
    \item We introduce learned blending schedules that anchor the transition to both motion clips at its boundaries, together with a contact planner whose foot contact schedule, injected through cross-attention adapters, keeps the feet on the ground in between.
    \item We conduct extensive evaluations on LaFAN1 and HumanML3D, which highlight the efficacy of our design and demonstrate that HumanBridge produces seamless transitions without foot sliding or floor penetration.
\end{itemize}

\begin{figure*}[ht]
    \centering
    \includegraphics[width=0.98\linewidth]{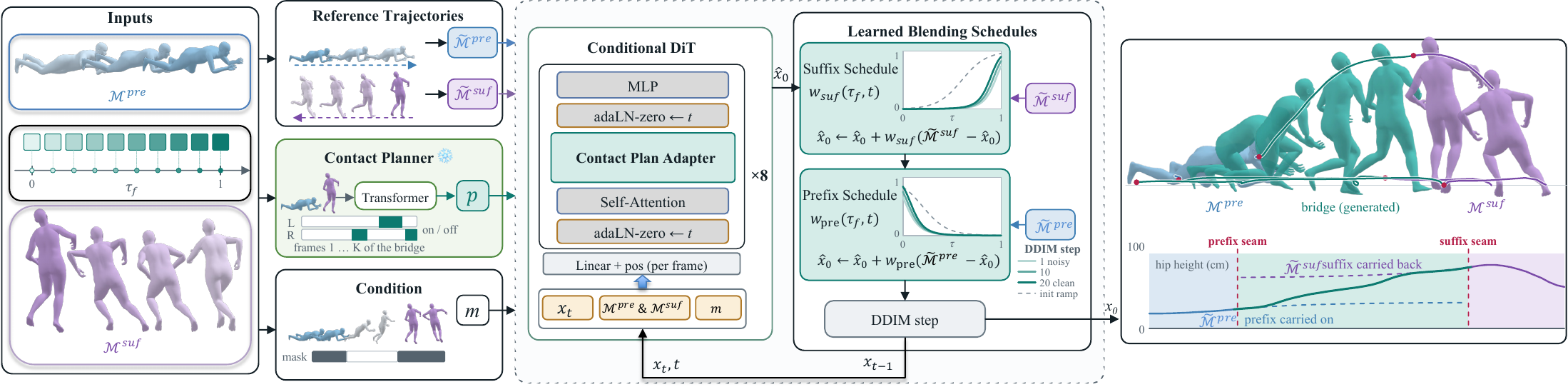}
    \vspace{-10pt}
    \caption{\textbf{Illustration of our HumanBridge framework.} Given a prefix clip and a suffix clip, HumanBridge generates the transition between them with a conditional diffusion model. A contact planner predicts when each foot should be on the ground, and the plan is injected into every DiT block through a cross-attention adapter. At each denoising step, two learned blending schedules guide the transition toward the prefix and suffix references near the two boundaries. The whole model is trained through the complete sampling chain without ground-truth transitions.} 
    \vspace{-10pt}
    \label{fig:framework}
\end{figure*}
\section{Related Work}
\noindent\textbf{Motion Transition and In-betweening.} Connecting two motion clips is a long-standing problem in character animation. Inertialization~\cite{bollo2018inertialization} decays the offset at the switch over time and is widely used in games, but it only smooths the offset between the two clips and does not synthesize new motion, so the feet slide when the two clips are in different phases of their steps. Learned in-betweening methods instead generate the frames between a context and a target keyframe. Harvey~\etal~\cite{harvey2020robust} propose a recurrent model with time-to-arrival embeddings together with the LaFAN1 dataset, Tang~\etal~\cite{tang2022real} learn a transition model for real-time character control, and transformer-based methods~\cite{qin2022motion,oreshkin2023motion,akhoundi2025silk} predict all missing frames in parallel to further improve the quality. Starke~\etal~\cite{starke2022deepphase,starke2023motion} further learn a periodic phase manifold of motion and use it to guide in-betweening, which, similar to our contact planner, models the stepping pattern explicitly. These methods are trained by masking out frames of continuous motion capture sequences, where the context and the target come from the same motion and the masked frames serve as ground truth. Different from them, we bridge two clips taken from different motions, for which no ground-truth transition exists, and therefore train our model without paired transition data.
Moreover, our contact planner predicts explicit foot contacts that continue the stepping pattern of the prefix clip into that of the suffix clip, and injects them into the denoiser.

\noindent\textbf{Diffusion-based Motion Generation.} Diffusion models~\cite{tevet2022human,rempe2026kimodo} have become the dominant approach for human motion generation, and they can fill in missing frames given the observed ones. CondMDI~\cite{cohan2024flexible} trains a diffusion model conditioned on randomly masked keyframes, and Scheduled Inpainting~\cite{agrawal2026interactive} edits motion by inpainting with a schedule over the denoising steps. For long motion composition, DoubleTake~\cite{shafir2024human} generates the transitions between text-driven segments with a second inpainting pass, and FlowMDM~\cite{barquero2024seamless} composes segments with blended positional encodings. Streaming and autoregressive methods~\cite{xiao2025motionstreamer,cai2026flooddiffusion,yu2026causal,zhao2026autoregressive,nazarenus2026actionplan,liu2026arms} generate each new segment from the motion produced so far, so that the action can change smoothly over time. Besides, guidance-based methods~\cite{karunratanakul2023guided,xie2024omnicontrol} steer the denoising process toward spatial targets such as the root trajectory or joint positions, and PhysDiff~\cite{yuan2023physdiff} projects the intermediate motion onto physically plausible states to reduce foot skating and floating. Although the guidance could in principle also target velocities, its targets and strength have to be specified and tuned by hand, and it does not decide when the feet should touch the ground. Overall, in these methods nothing ensures that the generated motion leaves the prefix and enters the suffix with matching velocities, which leads to visible jumps at the boundaries. Moreover, the streaming methods only condition on the past and never reach a fixed suffix clip. In contrast, HumanBridge guides each denoising step toward both clips with learned blending schedules, and keeps the feet on the ground in between with a predicted contact plan.

\section{Methodology}
\label{sec:method}
In this work, we aim to connect two existing motions by generating a seamless and natural transition. We propose HumanBridge, which takes the extrapolations of both prefix and suffix motion clips as priors to enforce velocity and pose continuity at both connection boundaries. Beyond that, we also adopt a foot contact planner to decide the foot contact schedule of the generated transition.

\subsection{Overview}
We build on a conditional diffusion transformer (DiT)~\cite{peebles2023scalable} with the conditioning scheme of CondMDI~\cite{cohan2024flexible}. The backbone $D_\theta$ takes a noisy motion window $\mathbf{x}_t$, an observation condition $\mathbf{c}$, and a mask $\mathbf{m}$, concatenated frame by frame, and directly predicts the clean motion $\hat{\mathbf{x}}_0$. Each frame $\mathbf{x}_f \in \mathbb{R}^{135}$ contains the 6D local rotations of 22 joints and the global root position. The mask is $1$ on the frames of both clips and $0$ on the transition, and the condition holds the two clips on their own frames and fills the transition with a linear interpolation between them, as in TST~\cite{qin2022motion}. We first pre-train the backbone as an in-betweening model on the training set, where a random gap is masked out of each motion and filled in from the surrounding frames, and add a foot contact loss~\cite{tevet2022human} in its last stage. This backbone then initializes our bridging model.

As illustrated in Figure~\ref{fig:framework}, given a prefix clip $\mathcal{M}^{\mathrm{pre}}$ with $N$ frames and a suffix clip $\mathcal{M}^{\mathrm{suf}}$ rigidly aligned to the last prefix frame, our goal is to generate a transition $\{\mathbf{x}_f\}_{f=N}^{N+K-1}$ of $K$ frames that connects seamlessly to both clips, which are played back unchanged. We use the following parametrization for the time axis, $\tau_f = (f - (N-1)) / (K+1)$, which goes from $0$ on the last prefix frame to $1$ on the first suffix frame. To this end, we first extrapolate the prefix forward and the suffix backward to obtain two reference trajectories, $\tilde{\mathcal{M}}^{\mathrm{pre}}$ and $\tilde{\mathcal{M}}^{\mathrm{suf}}$, while a contact planner $H_\vartheta$ predicts a foot contact schedule $\mathbf{p}$ for the transition. During DDIM~\cite{song2021denoising} sampling, a cross-attention adapter in each DiT block conditions every transition frame on its progress and planned contacts, and two learned blending schedules guide the clean-motion prediction toward the two references near the boundaries. Since no ground-truth transition exists, we train the whole model through its complete sampling chain, as detailed in the following sections.
\subsection{Learned Blending Schedules}
\label{sec:schedules}

A transition that connects seamlessly to both clips should initially continue the motion of the prefix and finally approach the suffix in the way the suffix itself was moving. However, conditioning the backbone on the observed frames only tells it where the clips are, not how fast they are moving, and the generated motion frequently jumps at the transition boundaries. To address this, we construct a reference trajectory for each clip and propose learned blending schedules that guide every denoising step toward these references near the boundaries.

\noindent\textbf{Boundary References.} As illustrated in Figure~\ref{fig:framework}, the prefix reference $\tilde{\mathcal{M}}^{\mathrm{pre}}$ extrapolates the last prefix frame forward with its final root velocity $\mathbf{v}$ and joint angular velocities $\boldsymbol{\omega}_j$, decayed by a factor $\lambda$ per frame. Specifically, the root position and the rotation of joint $j$ at frame $f$ are computed:
\begin{equation}
\begin{aligned}
  \tilde{\mathbf{p}}_f &= \mathbf{p}_{N-1} + S_n\,\mathbf{v}, \\
  \tilde{\mathbf{q}}_{j,f} &= \mathbf{q}_{j,N-1} \exp\bigl(S_n\,\boldsymbol{\omega}_j\bigr), 
\end{aligned}
\end{equation}
where $S_n = \sum_{i=1}^{n} \lambda^{i} = \frac{\lambda\,(1 - \lambda^{n})}{1 - \lambda}$ and $n = f - (N - 1)$, and $\mathbf{p}_{N-1}$ and $\mathbf{q}_{j,N-1}$ denote the root position and the local rotation of joint $j$ on the last prefix frame. In this way, the reference carries the momentum of the prefix into the transition and gradually comes to rest. Similarly, the suffix reference $\tilde{\mathcal{M}}^{\mathrm{suf}}$ extrapolates the first suffix frame backward in time with the initial velocities of the suffix, thus indicating where the character should be in order to arrive at the suffix at its own speed. Both references are computed from the motion clips alone and require no additional input.

\noindent\textbf{Blending Schedules.} Rather than imposing the references as hard constraints, we let the model learn how strongly to follow them at each frame and at each noise level. After the backbone predicts $\hat{\mathbf{x}}_0$ at diffusion step $t$, we blend the prediction on every transition frame toward the two references as follows, where $\tilde{\mathbf{x}}^{\mathrm{pre}}_f$ and $\tilde{\mathbf{x}}^{\mathrm{suf}}_f$ denote the features of $\tilde{\mathcal{M}}^{\mathrm{pre}}$ and $\tilde{\mathcal{M}}^{\mathrm{suf}}$ at frame $f$:
\begin{equation}
\begin{aligned}
  \hat{\mathbf{x}}_{0,f} &\leftarrow \hat{\mathbf{x}}_{0,f}
    + w_{\mathrm{suf}}(\tau_f, t)\,\bigl(\tilde{\mathbf{x}}^{\mathrm{suf}}_f - \hat{\mathbf{x}}_{0,f}\bigr), \\
  \hat{\mathbf{x}}_{0,f} &\leftarrow \hat{\mathbf{x}}_{0,f}
    + w_{\mathrm{pre}}(\tau_f, t)\,\bigl(\tilde{\mathbf{x}}^{\mathrm{pre}}_f - \hat{\mathbf{x}}_{0,f}\bigr),
\end{aligned}
\end{equation}
where $w_{\mathrm{pre}}$ and $w_{\mathrm{suf}}$ denote the prefix and suffix schedules. The blended prediction is then passed to the DDIM update, so that its effect carries over to all subsequent denoising steps. Each schedule is a sigmoid ramp over the transition progress, refined by a lightweight MLP:
\begin{equation}
\begin{aligned}
  w_{\mathrm{pre}}(\tau, t) &= \sigma\bigl(-\kappa\,(\tau - \mu_{\mathrm{pre}}) + \phi_{\mathrm{pre}}([\mathrm{PE}(\tau), s]) + \beta_{\mathrm{pre}}\bigr), \\
  w_{\mathrm{suf}}(\tau, t) &= \sigma\bigl(\phantom{-}\kappa\,(\tau - \mu_{\mathrm{suf}}) + \phi_{\mathrm{suf}}([\mathrm{PE}(\tau), s]) + \beta_{\mathrm{suf}}\bigr),
\end{aligned}
\end{equation}
where $\sigma$ is the sigmoid function, $\mathrm{PE}(\cdot)$ is a sinusoidal encoding, $s = t / (T - 1)$ is the normalized diffusion step, $\kappa$ controls the slope of the ramps, and $\mu_{\mathrm{pre}} < \mu_{\mathrm{suf}}$ are their centers. $\phi_{\mathrm{pre}}$ and $\phi_{\mathrm{suf}}$ are two-layer MLPs whose last layers are initialized to zero, and $\beta_{\mathrm{pre}}$ and $\beta_{\mathrm{suf}}$ are learnable scalars initialized to zero. Consequently, training starts from fixed ramps, in which the prefix schedule is strong at the beginning of the transition and fades out toward its middle, whereas the suffix schedule rises toward its end, and the MLPs learn how far each reference should reach into the transition at each noise level. The middle of the transition, where neither reference is meaningful, is thus left to the generative model. With about 2.2K parameters each, the schedules add a negligible cost to sampling.

\begin{figure}
    \centering
    \includegraphics[width=0.98\linewidth]{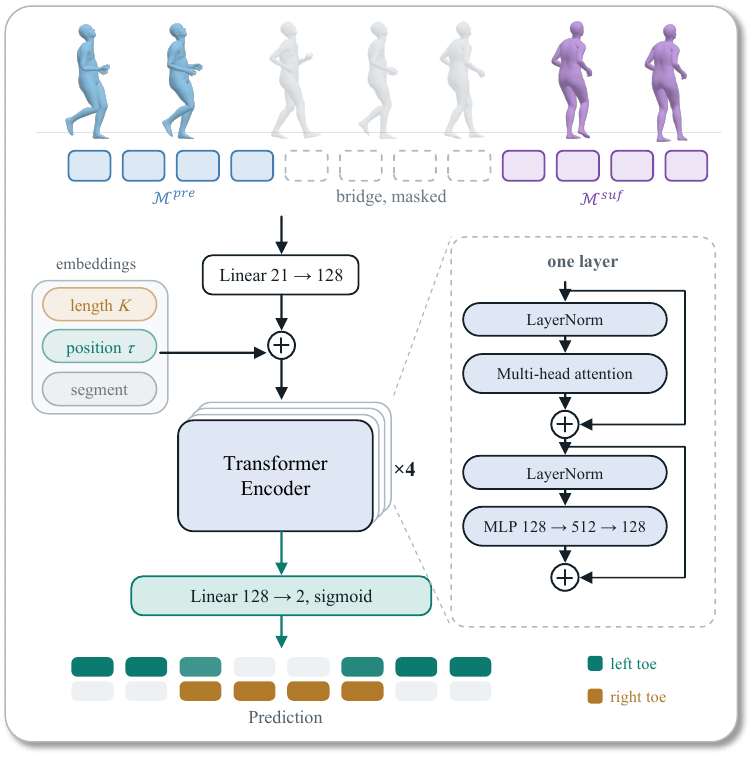}
    \vspace{-10pt}
    \caption{\textbf{The structure of the contact planner.} }
    \vspace{-10pt}
    \label{fig:contactplanner}
\end{figure}

\subsection{Contact-Aware Cross-Attention Adapters}
\label{sec:contacts}

The blending schedules determine how the transition leaves the prefix and enters the suffix, but not how the character steps in between. When the two clips are in different phases of their stepping patterns, a diffusion model without an explicit notion of contact tends to let the feet slide or to keep both of them off the ground for long stretches. To address this, we first predict when each foot should be on the ground with a contact planner, and then condition every layer of the denoiser on this plan through cross-attention adapters.

\noindent\textbf{Contact Planner.} As shown in Figure~\ref{fig:contactplanner}, the contact planner $H_\vartheta$ observes the last 10 frames of the prefix and the first 10 frames of the suffix. Each observed frame is described by a 21-dimensional feature $\mathbf{e}_f$, which contains the positions and velocities of the toes and the root, the contact states of the toes, and the height of the lower toe. Between the two clips, we insert $K$ masked tokens with $\mathbf{e}_f = \mathbf{0}$, and compute the input token of each frame as follows:
\begin{equation}
  \mathbf{z}_f = W_{\mathrm{in}}\,\mathbf{e}_f + E\bigl[\ell_f\bigr] + W_{\tau}\,\mathrm{PE}(\tau_f) + \mathbf{w}_K\,\frac{K}{K_{\max}},
\end{equation}
where $W_{\mathrm{in}}$ and $W_{\tau}$ are linear projections and $\mathbf{w}_K$ is a learnable vector. The segment label $\ell_f \in \{\mathrm{pre}, \mathrm{trans}, \mathrm{suf}\}$ selects an embedding from the learnable table $E$ to distinguish the observed frames from the masked ones, the progress $\tau_f$ locates each frame relative to the two boundaries, and the normalized length $K / K_{\max}$ with $K_{\max} = 45$ informs the planner of the time available for the transition. A transformer encoder followed by a sigmoid output layer predicts the contact probability of each toe on every transition frame, and thresholding these probabilities at $0.5$ yields the foot contact schedule $\mathbf{p} \in \{0, 1\}^{K \times 2}$. The planner is trained separately on continuous windows of real human motion with a binary cross-entropy loss, where a toe in the middle $K$ frames is labeled as in contact if it is within 5\,cm of the floor and moves horizontally by less than 1\,cm per frame. It is then kept frozen while the rest of the model is trained.


\noindent\textbf{Contact Plan Adapter.} Penalizing foot motion on the planned contact frames would only shape the training objective, leaving the plan invisible to the denoiser at inference. We therefore inject the plan into the denoiser through a cross-attention adapter in each DiT block, as illustrated in Figure~\ref{fig:framework}. Given the input $\mathbf{h}_f$ of the self-attention layer, the adapter computes
\begin{equation}
   \begin{aligned}
  \mathbf{q}_f &= W_q\,\mathrm{LN}_q(\mathbf{h}_f) + U_q\,\mathrm{PE}(\tau_f) + P\,\mathbf{p}_f, \\
  \mathbf{k}_j &= W_k\,\mathrm{LN}_{kv}(\mathbf{h}_j) + U_k\,\mathrm{PE}(\tau_j), \\
  \mathbf{v}_j &= W_v\,\mathrm{LN}_{kv}(\mathbf{h}_j),
\end{aligned} 
\end{equation}

where $W_q$, $W_k$, $W_v$, $U_q$, $U_k$ and $P$ are learnable projections, and $\mathbf{p}_f \in \{0, 1\}^2$ is the contact state of the two toes, taken from the planned schedule on the transition frames and measured from the clips on their own frames. With the contact state in its query, each transition frame looks for frames with a matching stepping pattern, while the progress in both queries and keys makes this search aware of the distance to the two boundaries. The adapter output
\begin{equation}
  \boldsymbol{\Delta}_f = g_f \tanh(\gamma)\, W_o\, \mathrm{Attn}\bigl(\mathbf{q}_f, \{\mathbf{k}_j\}, \{\mathbf{v}_j\}\bigr)
\end{equation}
is added to the self-attention output of the block, where $g_f \in \{0, 1\}$ restricts it to the transition frames, keeping the given clips unchanged. The learnable gate $\tanh(\gamma)$ bounds the adapter output, so that the plan adjusts the backbone features without disturbing them. Since $W_o$ and $P$ are initialized to zero, training starts exactly from the pre-trained backbone, which makes fine-tuning stable.

\subsection{Training Through the Sampling Chain}
\label{sec:fullchain}
\label{sec:training}

Existing in-betweening and diffusion methods are trained to reconstruct frames masked out of continuous motion. For bridging two existing clips, no ground-truth transition exists, so we instead optimize the model through the full denoising chain with losses defined directly on the generated transition, without any reconstruction target.

\noindent\textbf{Training Pairs.} To reproduce the mismatch at test time, the prefix and the suffix of each sample are drawn from two independent windows of the training set, so they typically disagree in pose, velocity and foot contacts. The transition length $K$ is randomly sampled for each batch.

\noindent\textbf{Full-Chain Training.} Without a clean target, the single-step denoising objective is not applicable. Instead, each training transition is generated exactly as at inference, by running the full DDIM sampling chain from pure noise with the adapters and blending schedules active at every step. Gradients are back-propagated through all steps, with activation checkpointing to save memory, and the losses are computed on the final output. In this way, the backbone, the adapters, and the blending schedules are jointly optimized for the final transition.

\noindent\textbf{Loss Functions.} All losses are computed on the generated motion and its global joint positions, with thresholds derived from real human motion in the training set, so that no loss favors a transition that is smoother, stiller or lower than real motion. Specifically, a reaching loss $\mathcal{L}_{\mathrm{reach}}$ pulls the end of the transition onto the suffix reference, a seam loss $\mathcal{L}_{\mathrm{seam}}$ penalizes excessive joint accelerations at the two boundaries, and a smoothness loss $\mathcal{L}_{\mathrm{smooth}}$ keeps the jerk within the range of real motion. For the contact with the ground, a skating loss $\mathcal{L}_{\mathrm{skate}}$ penalizes the horizontal motion of toes near the floor, a plan loss $\mathcal{L}_{\mathrm{plan}}$ keeps the planned contact toes still on the floor, a penetration loss $\mathcal{L}_{\mathrm{pen}}$ keeps the feet above the floor, and a hip loss $\mathcal{L}_{\mathrm{hip}}$ bounds the hip height. Finally, a naturalness loss $\mathcal{L}_{\mathrm{nat}}$ re-noises the output and denoises it with a frozen copy of the backbone, penalizing the change to keep the output on the learned motion manifold. The detailed definitions of all losses are provided in the supplementary material.
\begin{figure*}[ht]
    \centering
    \includegraphics[width=1\linewidth]{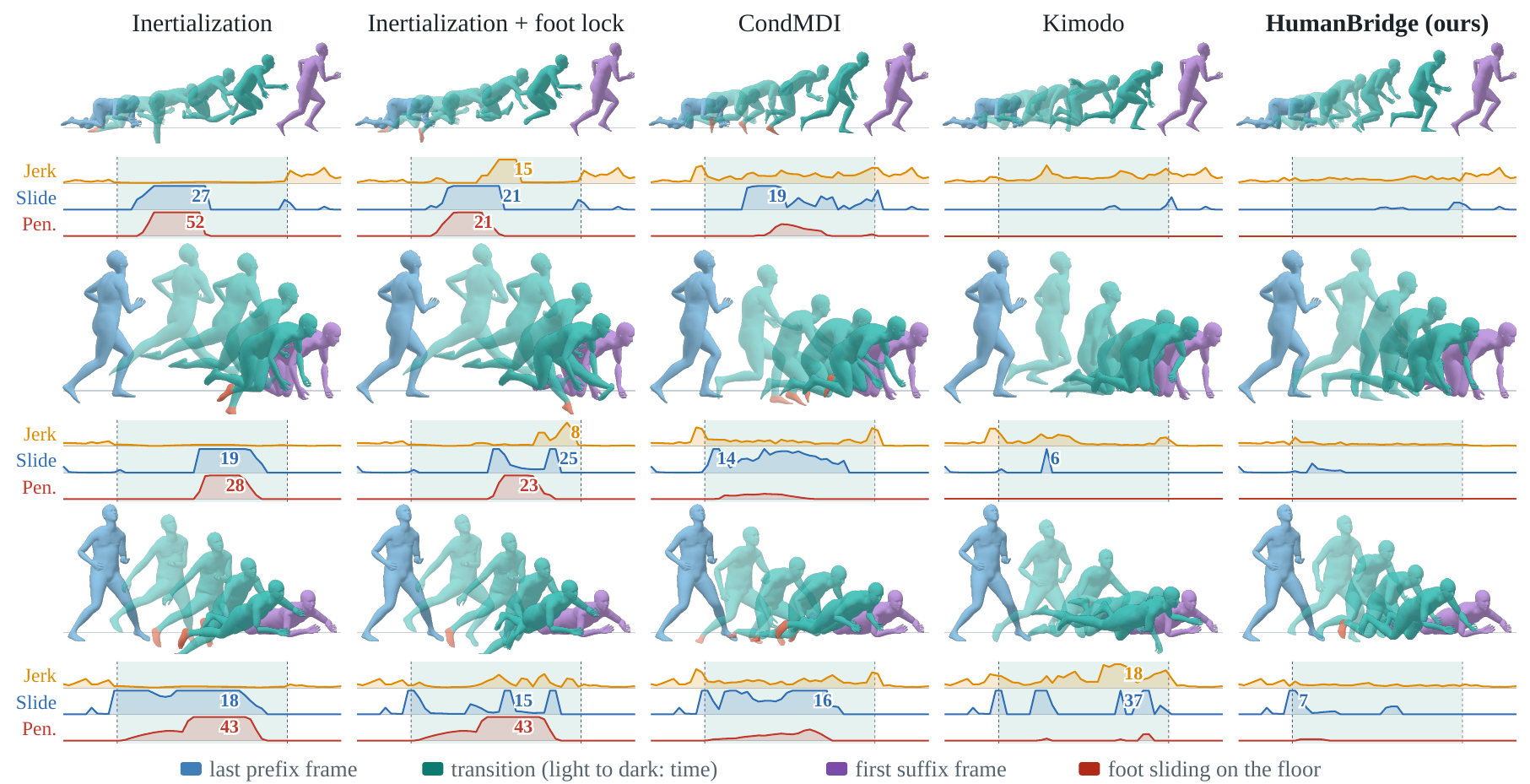}
    \vspace{-20pt}
    \caption{Qualitative comparison on three test pairs. Each panel shows the last prefix frame (blue), the transition (light to dark green) and the first suffix frame (purple), spread out horizontally for clarity, and red marks a sliding foot. The curves below show the jerk (0 to 8 times that of real motion), the foot sliding speed (0 to 6\,cm per frame) and the floor penetration (0 to 15\,cm) per frame, with the transition shaded. The baselines either jump at the boundaries or let the feet slide and sink into the floor, while HumanBridge keeps all three errors low.}
    \label{fig:qual}
    \vspace{-10pt}
\end{figure*}

\begin{figure}
    \centering
    \includegraphics[width=0.98\linewidth]{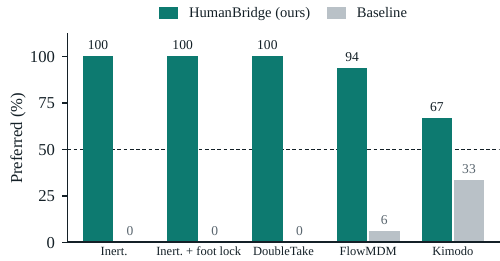}
    \vspace{-10pt}
    
    \caption{User study. For each baseline, the share of trials in which 30 participants judged our transition (green) or the baseline's (grey) more natural.}
    \label{fig:user_study}
    \vspace{-10pt}
    
\end{figure}

\section{Experiments}
\label{sec:experiments}

In this section, we present both quantitative and qualitative experiments to validate the effectiveness of HumanBridge. More results can be viewed in the supplementary material.

\subsection{Experimental Settings}
\label{sec:settings}
\noindent\textbf{Datasets.} We conduct our experiments on LaFAN1~\cite{harvey2020robust} and HumanML3D~\cite{guo2022generating}. For LaFAN1, we follow the standard split and use subjects 1 to 4 for training and subject 5 for testing. For HumanML3D, we use its training and test splits, and convert all motions to the LaFAN1 skeleton at 30 fps, so that both datasets share the same representation and the same metrics.


\noindent\textbf{Evaluation Metrics.} We evaluate two aspects. For the boundaries, the discontinuity rate (Disc.) is the fraction of pairs whose joint acceleration across either boundary exceeds three times the median of real motion. Bnd$^{\mathrm{pre}}$ and Bnd$^{\mathrm{suf}}$ are the mean joint accelerations across the two boundaries, and Jerk is the mean jerk within the transition, all relative to the median of real motion. PJ and AUJ are the peak jerk and the area under the jerk at the suffix boundary~\cite{barquero2024seamless}. For the ground contact, Skate is the mean horizontal speed (cm per frame) of the toes within 3\,cm of the floor, SkateR is the skating ratio of~\cite{cohan2024flexible}, Air is the fraction of transition frames with both toes more than 3\,cm above the floor, and Pen.\ is the fraction with the lower toe more than 2\,cm below it. Bnd, Jerk, and Air are best when closest to real motion rather than lowest, since a transition should be neither smoother nor stiller than real motion. Note that the discontinuity rate of real motion is not zero (15.7\% on LaFAN1), since real motion also contains sudden changes such as foot strikes. We therefore use Disc.\ to count clearly visible jumps, and use Bnd to check whether the boundaries are as smooth as real motion.

\begin{table*}[ht]
\centering
\scriptsize
\setlength{\tabcolsep}{3.2pt}
\caption{Quantitative comparison on LaFAN1 over 2219 test pairs. $\downarrow$: lower is better; $\rightarrow$: closer to real motion is better. Bnd$^{\mathrm{pre}}$, Bnd$^{\mathrm{suf}}$, and Jerk are given as multiples of the median of real motion, and Skate is in cm per frame. Foot lock pins the toes in contact with the floor with two-bone IK. The best generated result is shaded in red and the second best in orange; for $\rightarrow$ columns, the closest to real motion.}
\label{tab:lafan}
\vspace{-10pt}
\resizebox{\linewidth}{!}{%
\begin{tabular}{lcccccccccc}
\toprule
Method & Disc.\,(\%)$\downarrow$ & Bnd$^{\mathrm{pre}}\rightarrow$ & Bnd$^{\mathrm{suf}}\rightarrow$ & Jerk$\rightarrow$ & Skate$\downarrow$ & SkateR\,(\%)$\downarrow$ & Air\,(\%)$\rightarrow$ & Pen.\,(\%)$\downarrow$ & PJ$\downarrow$ & AUJ$\downarrow$ \\
\midrule
Real motion & 15.7 & 1.43 & 1.43 & 1.38 & 0.62 & 17.7 & 20.1 & 0.00 & 0.031 & 0.008 \\
\midrule
Inertialization~\cite{bollo2018inertialization} & \second{1.2} & 0.66 & \first{0.90} & 0.47 & 2.91 & 40.0 & 45.7 & 13.47 & \second{0.045} & 0.229 \\
Inertialization~\cite{bollo2018inertialization} + foot lock & 23.9 & 0.66 & 2.31 & 1.13 & 2.28 & 19.3 & 54.9 & 9.01 & 0.208 & 0.831 \\
TST~\cite{qin2022motion} & 54.4 & \first{1.97} & 3.86 & 1.90 & 2.34 & 42.2 & \first{20.2} & 9.17 & 0.078 & \second{0.182} \\
\midrule
DoubleTake~\cite{shafir2024human} & 100.0 & 19.99 & 25.01 & 6.79 & 7.34 & 35.9 & 65.8 & 15.72 & 0.954 & 2.036 \\
FlowMDM~\cite{barquero2024seamless} & 100.0 & 16.95 & 19.16 & 6.61 & 6.67 & 37.1 & 66.7 & 13.72 & 0.677 & 1.678 \\
CondMDI~\cite{cohan2024flexible} & 89.3 & 3.93 & 4.57 & 2.90 & 2.31 & 46.0 & \second{20.4} & 4.30 & 0.100 & 0.313 \\
Kimodo~\cite{rempe2026kimodo} & 56.6 & 2.84 & 2.77 & 2.52 & 1.96 & \first{12.3} & 67.4 & 1.48 & 0.184 & 0.673 \\
MotionBricks~\cite{wang2026motionbricks} & 89.0 & 4.26 & 3.89 & \first{1.39} & \second{1.85} & 27.3 & 44.7 & 4.14 & 0.184 & 0.390 \\
Sched.\ Inpainting~\cite{agrawal2026interactive} & 95.6 & 6.72 & 4.26 & 3.24 & 2.51 & 47.8 & \second{20.4} & 5.80 & 0.091 & 0.295 \\
ActionPlan~\cite{nazarenus2026actionplan} & 100.0 & 22.82 & 40.51 & 14.76 & 13.48 & 26.6 & 85.3 & \second{0.78} & 1.435 & 3.521 \\
\midrule
\textbf{HumanBridge (ours)} & \first{0.2} & \second{0.85} & \second{0.84} & \second{1.52} & \first{0.63} & \second{14.0} & \second{20.4} & \first{0.23} & \first{0.039} & \first{0.022} \\
\bottomrule
\end{tabular}
}
\vspace{-10pt}
\end{table*}

\begin{table*}[ht]
\centering
\scriptsize
\setlength{\tabcolsep}{3.2pt}

\caption{Quantitative comparison on HumanML3D over 6650 test pairs. \textit{Real motion via HumanML3D features} is real motion converted to HumanML3D features and back, the upper bound for DoubleTake, FlowMDM, and ActionPlan. $\downarrow$: lower is better; $\rightarrow$: closer to real motion is better. The best generated result is shaded in red and the second best in orange; for $\rightarrow$ columns, the closest to real motion.}
\vspace{-10pt}
\label{tab:hml3d}
\resizebox{\linewidth}{!}{%
\begin{tabular}{lcccccccccc}
\toprule
Method & Disc.\,(\%)$\downarrow$ & Bnd$^{\mathrm{pre}}\rightarrow$ & Bnd$^{\mathrm{suf}}\rightarrow$ & Jerk$\rightarrow$ & Skate$\downarrow$ & SkateR\,(\%)$\downarrow$ & Air\,(\%)$\rightarrow$ & Pen.\,(\%)$\downarrow$ & PJ$\downarrow$ & AUJ$\downarrow$ \\
\midrule
Real motion & 19.1 & 1.62 & 1.71 & 1.76 & 0.36 & 7.8 & 25.4 & 0.09 & 0.013 & 0.005 \\
Real motion via HumanML3D~\cite{guo2022generating} features & 60.2 & 4.16 & 4.40 & 2.28 & 0.41 & 7.9 & 26.0 & 0.10 & 0.035 & 0.072 \\
\midrule
Inertialization~\cite{bollo2018inertialization} & \second{6.6} & \second{0.82} & \first{1.14} & 0.99 & 1.54 & 30.8 & \second{29.1} & 9.71 & \second{0.020} & \second{0.076} \\
Inertialization~\cite{bollo2018inertialization} + foot lock & 42.5 & \second{0.82} & 3.71 & \second{2.47} & 1.16 & 17.6 & 35.9 & 8.13 & 0.163 & 0.614 \\
\midrule
DoubleTake~\cite{shafir2024human} & 100.0 & 39.79 & 57.09 & 16.02 & 5.95 & 15.4 & 79.6 & 6.57 & 0.801 & 1.724 \\
FlowMDM~\cite{barquero2024seamless} & 100.0 & 30.12 & 40.19 & 13.81 & 5.07 & 14.4 & 84.8 & 3.85 & 0.512 & 1.266 \\
CondMDI~\cite{cohan2024flexible} & 99.9 & 12.43 & 14.46 & 7.29 & \second{1.13} & 12.6 & 20.9 & 2.05 & 0.246 & 0.664 \\
Kimodo~\cite{rempe2026kimodo} & 97.4 & 5.24 & 5.23 & 5.32 & 1.20 & \first{8.5} & 73.5 & \second{0.42} & 0.091 & 0.520 \\
MotionBricks~\cite{wang2026motionbricks} & 100.0 & 7.09 & 6.52 & 2.55 & 1.17 & 18.9 & \first{26.8} & 3.90 & 0.084 & 0.187 \\
ActionPlan~\cite{nazarenus2026actionplan} & 100.0 & 40.10 & 67.28 & 32.72 & 5.54 & 21.1 & 69.9 & 1.58 & 0.966 & 2.442 \\
\midrule
\textbf{HumanBridge (ours)} & \first{4.0} & \first{1.04} & \second{1.02} & \first{2.20} & \first{0.46} & \second{11.6} & 17.1 & \first{0.05} & \first{0.018} & \first{0.015} \\
\bottomrule
\end{tabular}
}
\vspace{-10pt}
\end{table*}

\subsection{Comparison with State-of-the-Art Methods}
\label{sec:comparison}
In Table~\ref{tab:lafan} and Table~\ref{tab:hml3d}, we compare our method with inertialization~\cite{bollo2018inertialization}, TST~\cite{qin2022motion} and state-of-the-art generative methods (\ie, DoubleTake~\cite{shafir2024human}, FlowMDM~\cite{barquero2024seamless}, CondMDI~\cite{cohan2024flexible}, Kimodo~\cite{rempe2026kimodo}, MotionBricks~\cite{wang2026motionbricks}, Scheduled Inpainting~\cite{agrawal2026interactive} and ActionPlan~\cite{nazarenus2026actionplan}) on LaFAN1 and HumanML3D, where TST and Scheduled Inpainting are only evaluated on LaFAN1. HumanBridge achieves the best discontinuity rate, foot skating, floor penetration, PJ and AUJ on both datasets, and is second only to Kimodo in SkateR and to MotionBricks in jerk on LaFAN1.

\noindent\textbf{Quantitative Comparison.} As shown in Table~\ref{tab:lafan}, HumanBridge is the only method that keeps both the boundaries and the feet close to real motion, with a Disc of only 0.2\% and a Skate and Air that match real motion (0.63 vs.\ 0.62 and 20.4\% vs.\ 20.1\%). Inertialization also has a low Disc (1.2\%), but it only decays the offset between the clips and never generates new steps, so the feet slide (Skate 2.91) and sink into the floor (Pen 13.47\% vs.\ 0.23\%). Pinning the feet with IK lowers Skate to 2.28 but raises Disc to 23.9\%, since the locked feet must be released suddenly to reach the suffix. The learned and diffusion methods generate new steps but jump at the boundaries (Disc of at least 54.4\%), since they know where the clips are but not how fast they move. Kimodo has a lower SkateR than ours only because both of its feet are in the air in 67.4\% of the frames. MotionBricks keeps the jerk close to real motion (1.39 vs.\ 1.38), but it still jumps at the boundaries (Disc of 89.0\%) and keeps both feet in the air in 44.7\% of the frames. CondMDI and Scheduled Inpainting share our backbone but still reach a Disc of 89.3\% and 95.6\%, so the gain comes from our bridging design rather than the backbone. Moreover, our boundaries are not over-smoothed. Our Jerk stays close to real motion (1.52 vs.\ 1.38, while inertialization has 0.47), and our AUJ is ten times lower (0.022 vs.\ 0.229). On HumanML3D (Table~\ref{tab:hml3d}), HumanBridge again has the lowest Disc (4.0\% vs.\ 6.6\% for inertialization) and a 3.3 times lower Skate (0.46 vs.\ 1.54). On one A100 GPU, HumanBridge generates a 30-frame transition in 0.23\,s (3.6\,ms per transition in batches of 64) vs.\ 0.18\,s for CondMDI with the same backbone, so the blending schedules, the contact planner and the adapters add only about 30\% to the sampling cost.

\noindent\textbf{Qualitative Comparison.} Additionally, we visualize some comparison results in Figure~\ref{fig:qual}, with the jerk, foot sliding and floor penetration of every frame shown below each result. As shown, inertialization keeps the jerk low, but since it only blends the offset between the clips, its feet slide up to 27\,cm per frame and sink up to 52\,cm into the floor. Pinning the feet with IK does not prevent the penetration (up to 43\,cm) and adds jerk peaks of up to 15 times that of real motion, since the locked feet are released suddenly. CondMDI shows jerk peaks right at the boundaries and keeps sliding during the transition, and Kimodo reaches a jerk of 18 times that of real motion and a sliding speed of 37\,cm per frame. In contrast, the jerk of HumanBridge has no peak at either boundary and stays at the level of the two clips, and its feet barely slide and sink at most 1\,cm into the floor. This shows that our method connects to both clips without a jump while keeping the feet on the ground. 

\noindent\textbf{User Study.} We conduct a user study with 30 participants, who watch our transition and that of one baseline on the same pair side by side and choose the more natural one. As shown in Figure~\ref{fig:user_study}, participants prefer HumanBridge in 91.5\% of the trials and flag far fewer of our transitions as glitchy (12.2\% vs.\ 79.3\%). Kimodo is the closest competitor (67\% vs.\ 33\%), since it also generates new steps, but its transitions are still flagged more often (38.9\% vs.\ 16.7\%). 

\subsection{Ablation Study}

\begin{table}[t]
\centering
\setlength{\tabcolsep}{3.2pt}
\caption{Ablation study on LaFAN1. Each row changes one component of the full model, except the first, which removes both the blending schedules and the adapters. $\downarrow$: lower is better; $\rightarrow$: closer to real motion is better.}
\vspace{-10pt}
\label{tab:ablation}
\resizebox{\linewidth}{!}{%
\begin{tabular}{lccccccc}
\toprule
Method & Disc.\,(\%)$\downarrow$ & Bnd$^{\mathrm{pre}}\rightarrow$ & Bnd$^{\mathrm{suf}}\rightarrow$ & Jerk$\rightarrow$ & Skate$\downarrow$ & Air\,(\%)$\rightarrow$ & Pen.\,(\%)$\downarrow$ \\
\midrule
Real motion & 15.7 & 1.43 & 1.43 & 1.38 & 0.62 & 20.1 & 0.00 \\
\midrule
Losses only (w/o schedules, adapter) & 5.0 & 1.24 & 1.49 & 1.88 & 1.09 & 14.2 & 0.16 \\
\midrule
w/o prefix schedule & 1.0 & 1.22 & 0.88 & 1.71 & 0.79 & 18.5 & 0.25 \\
w/o suffix schedule & 0.1 & 0.83 & 0.79 & 1.33 & 0.94 & 19.2 & 0.88 \\
\midrule
w/o contact plan & 0.1 & 0.82 & 0.82 & 1.32 & 0.87 & 86.0 & 0.08 \\
Single-step training & 0.7 & 0.82 & 1.02 & 1.38 & 1.56 & 23.3 & 1.86 \\
w/o $\mathcal{L}_{\mathrm{skate}}$ & 0.2 & 0.86 & 0.84 & 1.54 & 1.93 & 12.6 & 0.29 \\
w/o $\mathcal{L}_{\mathrm{pen}}$, $\mathcal{L}_{\mathrm{hip}}$ floor & 0.1 & 0.82 & 0.82 & 1.46 & 0.64 & 18.8 & 13.42 \\
\midrule
\textbf{Full model} & 0.2 & 0.85 & 0.84 & 1.52 & 0.63 & 20.4 & 0.23 \\
\bottomrule
\end{tabular}}
\vspace{-10pt}
\end{table}
In this section, we conduct ablation studies to verify the effectiveness of each component of HumanBridge. Each variant removes one component of the full model, and is trained and evaluated on LaFAN1. The quantitative results are shown in Table~\ref{tab:ablation}, and the qualitative results are given in the supplementary material.

\noindent\textbf{Losses Only.} To check whether the gain comes from our losses rather than from the two modules, we train the backbone through the full sampling chain with all losses, but without the blending schedules and the adapters. The losses alone already lower the discontinuity rate of the backbone (5.0\% vs.\ 89.3\% for CondMDI), but the transitions still jump at the boundaries 25 times more often than with the full model (5.0\% vs.\ 0.2\%), and the jerk moves further away from real motion (1.88 vs.\ 1.52, compared with 1.38). Foot skating is also 1.7 times higher (1.09 vs.\ 0.63), and the feet leave the ground less often than in real motion (14.2\% vs.\ 20.1\%), so they drag on the floor. This shows that the losses tell the model what a good transition is, while the blending schedules and the contact plan let the denoiser actually produce it.

\noindent\textbf{Blending Schedules.} The prefix schedule makes the transition leave the prefix clip with the right velocity. Without it, the discontinuity rate increases five times (1.0\% vs.\ 0.2\%), and the jerk moves away from real motion (1.71 vs.\ 1.52, compared with 1.38), since the wrong initial velocity has to be corrected inside the transition. The suffix schedule instead gives the model a target to approach. Without it, the reaching loss still pulls the last frames onto the suffix clip, so the discontinuity rate does not increase. However, the model closes the remaining distance by sliding the feet, which increases foot skating from 0.63 to 0.94.

\noindent\textbf{Contact Plan.} Without the contact plan, both feet are off the ground in 86.0\% of the frames (vs.\ 20.1\% in real motion), while foot skating rises only slightly (0.87 vs.\ 0.63) and floor penetration even drops (0.08\% vs.\ 0.23\%). We think the reason is that the skating and penetration losses only act on feet near the floor, so the model can satisfy both of them by simply lifting both feet. In the qualitative results of the supplementary material, the character floats for 24 to 28 consecutive frames with the lower foot 8 to 9\,cm above the floor, whereas the full model keeps a foot on the ground in 24 to 30 of the 30 frames. Moreover, the boundaries are not affected (0.1\%), which suggests that the contact plan and the blending schedules address two separate problems.

\noindent\textbf{Training and Losses.} Single-step training more than doubles foot skating (1.56) and increases the discontinuity rate to 0.7\%, since the losses on a one-step estimate do not reflect the transition produced after all 20 sampling steps. Removing the skating loss triples foot skating (1.93), and the feet leave the ground less often than in real motion (12.6\% vs.\ 20.1\%), which means that they drag on the floor. Together with the result without the contact plan (86.0\%), this suggests that the skating loss and the contact plan balance each other. The skating loss makes the feet lift and the plan makes them land, so the air ratio of the full model is close to that of real motion (20.4\% vs.\ 20.1\%). Finally, removing the penetration loss and the lower bound of the hip loss raises floor penetration from 0.23\% to 13.42\%, with the toes sinking 4.8 to 6.4\,cm at foot strikes, while the other metrics barely change.

\section{Conclusion}
\label{sec:conclusion}

In this work, we introduce HumanBridge, a diffusion-based framework that bridges two existing motion clips with seamless transitions and high-quality foot motion while playing both clips back unchanged. Our method uses two learned blending schedules to guide each denoising step toward the prefix and suffix clips near the transition boundaries, so that the transition connects to both clips with continuous velocity. Further, a contact planner predicts when each foot should be on the ground, and cross-attention adapters inject this plan into every block of the denoiser. Since no ground-truth transition exists, we train the whole model through its complete sampling chain. Extensive experiments on LaFAN1 and HumanML3D demonstrate that HumanBridge nearly eliminates the jumps at the boundaries and keeps the feet on the ground as in real human motion, achieving superior results compared to existing methods. Ablation studies verify the effectiveness of the blending schedules in connecting the two clips and of the contact plan in preventing the feet from sliding, floating and sinking into the floor.

\section*{LLM Usage Statement}
We used a large language model (LLM) as a writing and coding assistant in this work.
For writing, it helped polish the language of drafts written by the authors, such as fixing grammar and shortening sentences.
For coding, it helped write and debug parts of the training, evaluation, and plotting scripts.
The LLM did not propose the research idea, the method, or the experiment design.
All code was run, checked, and verified by the authors, and every number in the paper comes from our own experiments.
The authors take full responsibility for the content of this paper.

{
    \small
    \bibliographystyle{ieeenat_fullname}
    \bibliography{main}

@String(CVPR= {IEEE Conf. Comput. Vis. Pattern Recog.})

@String(ICCV= {Int. Conf. Comput. Vis.})

@String(TOG= {ACM Trans. Graph.})

@String(ICLR = {Int. Conf. Learn. Represent.})

@String(CVPRW= {IEEE Conf. Comput. Vis. Pattern Recog. Worksh.})

@String(CVPR  = {CVPR})

@String(ICCV  = {ICCV})

@String(TOG   = {ACM TOG})

@String(ICLR  = {ICLR})

@String(CVPRW= {CVPRW})

@article{zhao2026autoregressive,
  title={Autoregressive diffusion with hybrid representation for interactive human motion generation},
  author={Zhao, Kaifeng and Petrovich, Mathis and Zhang, Haotian and Wang, Tingwu and Tang, Siyu and Rempe, Davis},
  journal={ACM Transactions on Graphics (TOG)},
  volume={45},
  number={4},
  pages={1--14},
  year={2026},
  publisher={ACM, New York, NY, USA}
}

@article{rempe2026kimodo,
  title={Kimodo: Scaling controllable human motion generation},
  author={Rempe, Davis and Petrovich, Mathis and Yuan, Ye and Zhang, Haotian and Peng, Xue Bin and Jiang, Yifeng and Wang, Tingwu and Iqbal, Umar and Minor, David and de Ruyter, Michael and others},
  journal={arXiv preprint arXiv:2603.15546},
  year={2026}
}

@article{agrawal2026interactive,
  title={Interactive Generative Motion Editing via Scheduled Inpainting},
  author={Agrawal, Dhruv and Borer, Dominik and V{\"o}geli, Luca and Sumner, Robert and Guay, Martin and Buhmann, Jakob},
  journal={arXiv preprint arXiv:2607.29133},
  year={2026}
}

@inproceedings{nazarenus2026actionplan,
  title={Actionplan: Future-aware streaming motion synthesis via frame-level action planning},
  author={Nazarenus, Eric and Li, Chuqiao and He, Yannan and Xie, Xianghui and Lenssen, Eric Jan and Pons-Moll, Gerard},
  booktitle={European Conference on Computer Vision},
  pages={553--573},
  year={2026},
  organization={Springer}
}

@inproceedings{xiao2025motionstreamer,
  title={Motionstreamer: Streaming motion generation via diffusion-based autoregressive model in causal latent space},
  author={Xiao, Lixing and Lu, Shunlin and Pi, Huaijin and Fan, Ke and Pan, Liang and Zhou, Yueer and Feng, Ziyong and Zhou, Xiaowei and Peng, Sida and Wang, Jingbo},
  booktitle={2025 IEEE/CVF International Conference on Computer Vision (ICCV)},
  pages={10086--10096},
  year={2025},
  organization={IEEE}
}

@inproceedings{liu2026arms,
  title={ARMS: Anchor--Relational Motion Streaming for Seamless Solo-Social Motion Transitions},
  author={Liu, Huakun and Yu, Qing and Fujiwara, Kent and Uchiyama, Hideaki and Kiyokawa, Kiyoshi},
  booktitle={European Conference on Computer Vision},
  pages={509--527},
  year={2026},
  organization={Springer}
}

@inproceedings{cai2026flooddiffusion,
  title={Flooddiffusion: Tailored diffusion forcing for streaming motion generation},
  author={Cai, Yiyi and Wu, Yuhan and Li, Kunhang and Zhou, You and Zheng, Bo and Liu, Haiyang},
  booktitle={Proceedings of the IEEE/CVF Conference on Computer Vision and Pattern Recognition},
  pages={2295--2304},
  year={2026}
}

@article{yu2026causal,
  title={Causal motion diffusion models for autoregressive motion generation},
  author={Yu, Qing and Watanabe, Akihisa and Fujiwara, Kent},
  journal={arXiv preprint arXiv:2602.22594},
  year={2026}
}

@article{harvey2020robust,
  title={Robust motion in-betweening},
  author={Harvey, F{\'e}lix G and Yurick, Mike and Nowrouzezahrai, Derek and Pal, Christopher},
  journal={ACM Transactions on Graphics (TOG)},
  volume={39},
  number={4},
  pages={60--1},
  year={2020},
  publisher={ACM, New York, NY, USA}
}

@article{tang2022real,
  title={Real-time controllable motion transition for characters},
  author={Tang, Xiangjun and Wang, He and Hu, Bo and Gong, Xu and Yi, Ruifan and Kou, Qilong and Jin, Xiaogang},
  journal={ACM Transactions on Graphics (TOG)},
  volume={41},
  number={4},
  pages={1--10},
  year={2022},
  publisher={ACM, New York, NY, USA}
}

@inproceedings{cohan2024flexible,
  title={Flexible motion in-betweening with diffusion models},
  author={Cohan, Setareh and Tevet, Guy and Reda, Daniele and Peng, Xue Bin and Van de Panne, Michiel},
  booktitle={ACM SIGGRAPH 2024 conference papers},
  pages={1--9},
  year={2024}
}

@article{qin2022motion,
  title={Motion in-betweening via two-stage transformers},
  author={Qin, Jia and Zheng, Youyi and Zhou, Kun},
  journal={ACM Transactions on Graphics (TOG)},
  volume={41},
  number={6},
  pages={1--16},
  year={2022},
  publisher={ACM, New York, NY, USA}
}

@inproceedings{akhoundi2025silk,
  title={SILK: Smooth InterpoLation Framework for Motion in-Betweening a Simplified Computational Approach},
  author={Akhoundi, Elly and Ling, Hung Yu and Deshmukh, Anup and Butepage, Judith},
  booktitle={2025 IEEE/CVF Conference on Computer Vision and Pattern Recognition Workshops (CVPRW)},
  pages={2891--2900},
  year={2025},
  organization={IEEE}
}

@article{oreshkin2023motion,
  title={Motion In-Betweening via Deep {$\Delta$}-Interpolator},
  author={Oreshkin, Boris N and Valkanas, Antonios and Harvey, F{\'e}lix G and M{\'e}nard, Louis-Simon and Bocquelet, Florent and Coates, Mark J},
  journal={IEEE Transactions on Visualization and Computer Graphics},
  volume={30},
  number={8},
  pages={5693--5704},
  year={2023},
  publisher={IEEE}
}

@inproceedings{bollo2018inertialization,
  title={Inertialization: High-performance animation transitions in’gears of war’},
  author={Bollo, David},
  booktitle={Proc. of GDC},
  volume={2016},
  year={2018}
}

@incollection{kovar2023motion,
  title={Motion graphs},
  author={Kovar, Lucas and Gleicher, Michael and Pighin, Fr{\'e}d{\'e}ric},
  booktitle={Seminal Graphics Papers: Pushing the Boundaries, Volume 2},
  pages={723--732},
  year={2023}
}

@article{tevet2022human,
  title={Human motion diffusion model},
  author={Tevet, Guy and Raab, Sigal and Gordon, Brian and Shafir, Yonatan and Cohen-Or, Daniel and Bermano, Amit H},
  journal={arXiv preprint arXiv:2209.14916},
  year={2022}
}

@inproceedings{shafir2024human,
  title={Human motion diffusion as a generative prior},
  author={Shafir, Yonatan and Tevet, Guy and Kapon, Roy and Bermano, Amit},
  booktitle={International Conference on Learning Representations},
  volume={2024},
  pages={8717--8733},
  year={2024}
}

@inproceedings{barquero2024seamless,
  title={Seamless human motion composition with blended positional encodings},
  author={Barquero, German and Escalera, Sergio and Palmero, Cristina},
  booktitle={2024 IEEE/CVF Conference on Computer Vision and Pattern Recognition (CVPR)},
  pages={457--469},
  year={2024},
  organization={IEEE}
}

@inproceedings{guo2022generating,
  title={Generating diverse and natural 3d human motions from text},
  author={Guo, Chuan and Zou, Shihao and Zuo, Xinxin and Wang, Sen and Ji, Wei and Li, Xingyu and Cheng, Li},
  booktitle={2022 IEEE/CVF Conference on Computer Vision and Pattern Recognition (CVPR)},
  pages={5142--5151},
  year={2022},
  organization={IEEE}
}

@inproceedings{tang2023rsmt,
  title={Rsmt: Real-time stylized motion transition for characters},
  author={Tang, Xiangjun and Wu, Linjun and Wang, He and Hu, Bo and Gong, Xu and Liao, Yuchen and Li, Songnan and Kou, Qilong and Jin, Xiaogang},
  booktitle={ACM SIGGRAPH 2023 Conference Proceedings},
  pages={1--10},
  year={2023}
}

@inproceedings{peebles2023scalable,
  title={Scalable diffusion models with transformers},
  author={Peebles, William and Xie, Saining},
  booktitle={2023 IEEE/CVF International Conference on Computer Vision (ICCV)},
  pages={4172--4182},
  year={2023},
  organization={IEEE}
}

@article{song2021denoising,
  title={Denoising diffusion implicit models},
  author={Song, Jiaming and Meng, Chenlin and Ermon, Stefano},
  journal={ICLR},
  year={2020}
}

@inproceedings{karunratanakul2023guided,
  title={Guided motion diffusion for controllable human motion synthesis},
  author={Karunratanakul, Korrawe and Preechakul, Konpat and Suwajanakorn, Supasorn and Tang, Siyu},
  booktitle={2023 IEEE/CVF International Conference on Computer Vision (ICCV)},
  pages={2151--2162},
  year={2023},
  organization={IEEE}
}

@inproceedings{xie2024omnicontrol,
  title={Omnicontrol: Control any joint at any time for human motion generation},
  author={Xie, Yiming and Jampani, Varun and Zhong, Lei and Sun, Deqing and Jiang, Huaizu},
  booktitle={International Conference on Learning Representations},
  volume={2024},
  pages={28176--28194},
  year={2024}
}

@inproceedings{yuan2023physdiff,
  title={Physdiff: Physics-guided human motion diffusion model},
  author={Yuan, Ye and Song, Jiaming and Iqbal, Umar and Vahdat, Arash and Kautz, Jan},
  booktitle={2023 IEEE/CVF International Conference on Computer Vision (ICCV)},
  pages={15964--15975},
  year={2023},
  organization={IEEE}
}

@article{starke2023motion,
  title={Motion in-betweening with phase manifolds},
  author={Starke, Paul and Starke, Sebastian and Komura, Taku and Steinicke, Frank},
  journal={Proceedings of the ACM on Computer Graphics and Interactive Techniques},
  volume={6},
  number={3},
  pages={1--17},
  year={2023},
  publisher={ACM, New York, NY, USA}
}

@article{starke2022deepphase,
  title={Deepphase: Periodic autoencoders for learning motion phase manifolds},
  author={Starke, Sebastian and Mason, Ian and Komura, Taku},
  journal={ACM Transactions on Graphics (ToG)},
  volume={41},
  number={4},
  pages={1--13},
  year={2022},
  publisher={ACM, New York, NY, USA}
}

@article{wang2026motionbricks,
  title={MotionBricks: Scalable Real-Time Motions with Modular Latent Generative Model and Smart Primitives},
  author={Wang, Tingwu and Dionne, Olivier and De Ruyter, Michael and Minor, David and Rempe, Davis and Zhao, Kaifeng and Petrovich, Mathis and Yuan, Ye and Li, Chenran and Luo, Zhengyi and others},
  journal={ACM Transactions on Graphics (TOG)},
  volume={45},
  number={4},
  pages={1--22},
  year={2026},
  publisher={ACM, New York, NY, USA}
}
}

\clearpage
\appendix
\newcounter{mainpage}
\setcounter{mainpage}{\value{page}}
\renewcommand{\maketitlesupplementary}{\setcounter{page}{\value{mainpage}}%
  \twocolumn[\centering\Large\textbf{Supplementary Material}\\\vspace{1.0em}]}
\clearpage
\setcounter{page}{1}
\maketitlesupplementary

This supplementary material provides the experimental details (Section~\ref{sec:suppl_details}), the definitions of all losses (Section~\ref{sec:suppl_loss}), the baseline results without post-processing (Section~\ref{sec:suppl_raw}), the results for a longer transition (Section~\ref{sec:suppl_k}), the accuracy of the contact planner (Section~\ref{sec:suppl_plan}), additional qualitative results (Section~\ref{sec:suppl_qual}) and the limitations of our method (Section~\ref{sec:suppl_limit}).

\section{Experimental Details}
\label{sec:suppl_details}

\paragraph{Bridging Protocol.} Since no benchmark exists for bridging two existing clips, we build one from the test set of each dataset. Specifically, we cut the test motions into windows of 105 frames with a stride of 40 frames, which gives 2219 windows on LaFAN1 and 6650 windows on HumanML3D. For the $i$-th window, the prefix clip is its first $N = 30$ frames, and the suffix clip is taken from the window halfway across the test set, rigidly aligned to the last prefix frame in ground position and heading. In this way, the two clips come from different motions and disagree in pose, velocity and foot contacts, as in practice. Each method generates a transition of $K$ frames, after which the suffix clip is played back unchanged for 30 frames. We report $K = 30$ (1\,s) in the main paper, and $K = 45$ (1.5\,s) in Section~\ref{sec:suppl_k}. Since no ground-truth transition exists, per-frame errors such as L2P and NPSS cannot be computed. Instead, we compare every method with real human motion, \ie, the same prefix followed by its own continuation in the original sequence, evaluated at the same frames.

\paragraph{Baselines.} We compare HumanBridge with transition, in-betweening and diffusion methods. For transitions, we use inertialization~\cite{bollo2018inertialization}, which is widely used in game engines and decays the offset between the two clips within the $K$ frames. We also add a stronger variant that further pins the toes in contact to the floor with two-bone IK (foot lock), as a common game-engine pipeline. For in-betweening, we use TST~\cite{qin2022motion} with its official LaFAN1 weights, taking the last 10 prefix frames and the first 2 suffix frames as input. Since TST is only trained on LaFAN1, it is not used on HumanML3D. For diffusion methods, we use Kimodo~\cite{rempe2026kimodo} with its official weights, taking the prefix and the first 30 suffix frames as full-body constraints. For MotionBricks~\cite{wang2026motionbricks}, we use its official model, which generates the motion from the last four prefix frames to the first four suffix frames and corrects both endpoints to match them exactly. Since it only generates transitions of $4n - 6$ frames, we use the length closest to $K$ and resample the result in time to $K$ frames, which changes the playback speed of the transition by at most 6\%. The motions of both datasets are retargeted to its skeleton by joint positions, and the results are mapped back in the same way. We use ActionPlan~\cite{nazarenus2026actionplan}, FlowMDM~\cite{barquero2024seamless} and DoubleTake~\cite{shafir2024human} with their official HumanML3D weights, inpainting the transition between the two clips. Since these three methods generate the root as velocities, the generated suffix drifts away from the given one, so we report them with a post-processing (PP) that rigidly aligns the generated suffix to the given one and blends the offset into the transition. This improves most of their metrics, and the raw results are given in Section~\ref{sec:suppl_raw}. Finally, for a controlled comparison, we use our own backbone before bridging training, which takes both clips as observed frames as in CondMDI~\cite{cohan2024flexible} and is trained on the same data (denoted as CondMDI), and Scheduled Inpainting~\cite{agrawal2026interactive} on the same backbone.

\paragraph{Implementation Details.} All motions are processed at 30 fps with $N = 30$ visible prefix frames. The backbone consists of 8 DiT blocks of width 512 with adaLN-zero conditioning and a cosine noise schedule with $T = 1000$ steps. It is pre-trained for 150K steps on gaps of 5 to 45 frames, and fine-tuned with a foot contact loss for 30K steps. The contact planner is a 4-layer transformer encoder of width 128 trained for 20K steps. The adapters use 4 heads of width 32 with $\gamma$ initialized to 0.1, and add 2.19M parameters to the 39.1M of the backbone. For the blending schedules, we set $\lambda = 0.9$, $\kappa = 8$, $\mu_{\mathrm{pre}} = 0.2$ and $\mu_{\mathrm{suf}} = 0.6$. HumanBridge is trained for 8K steps with a batch size of 64 and $K \in [5, 45]$, using 20 DDIM steps and learning rates of $3 \times 10^{-5}$, $10^{-4}$ and $10^{-3}$ for the backbone, the adapters and the blending schedules, respectively. All statistics used by the losses are computed on the training set only. In the output, the frames of both clips are always replaced by the given ones, so both clips are played back exactly. On HumanML3D, we retrain the entire pipeline on its training set.

\section{Loss Functions}
\label{sec:suppl_loss}
In this section, we give the definitions of the losses used to train HumanBridge. All of them are computed on the final output of the sampling chain. Let $\mathbf{g}_{f,j} \in \mathbb{R}^3$ denote the global position (in cm) of joint $j$ at frame $f$ of the generated motion, with $J = 22$ joints and $y_{f,j}$ its height, and let $\mathcal{T} = \{N, \dots, N+K-1\}$ denote the transition frames. We measure the mean joint acceleration and the mean joint jerk of a frame with finite differences,
\begin{equation}
\begin{aligned}
  a_f &= \frac{1}{J} \sum_{j} \bigl\| \mathbf{g}_{f+1,j} - 2\,\mathbf{g}_{f,j} + \mathbf{g}_{f-1,j} \bigr\|, \\
  r_f &= \frac{1}{J} \sum_{j} \bigl\| \mathbf{g}_{f+3,j} - 3\,\mathbf{g}_{f+2,j} + 3\,\mathbf{g}_{f+1,j} - \mathbf{g}_{f,j} \bigr\|,
\end{aligned}
\end{equation}
and denote by $\bar{a}$ and $\bar{r}$ their medians over real motion in the training set. The floor height $h_0$ is the 5th percentile of the toe heights in the training set.

\noindent\textbf{Reaching Loss.} The reaching loss pulls the last three transition frames onto the suffix reference $\tilde{\mathcal{M}}^{\mathrm{suf}}$,
\begin{equation}
  \mathcal{L}_{\mathrm{reach}} = \frac{1}{9J} \sum_{f=N+K-3}^{N+K-1} \sum_{j} \Bigl\| \bigl(\mathbf{g}_{f,j} - \tilde{\mathbf{g}}^{\mathrm{suf}}_{f,j}\bigr) \oslash \boldsymbol{\sigma}_j \Bigr\|^2,
\end{equation}
where $\tilde{\mathbf{g}}^{\mathrm{suf}}_{f,j}$ is the joint position of the suffix reference, $\oslash$ is the element-wise division, and $\boldsymbol{\sigma}_j$ is the standard deviation of the position of joint $j$ in the training set.

\noindent\textbf{Seam Loss.} The seam loss penalizes the joint acceleration on the two frames around each boundary, $\mathcal{B} = \{N-1, N, N+K-1, N+K\}$, but only above the level of real motion,
\begin{equation}
  \mathcal{L}_{\mathrm{seam}} = \frac{1}{4\,\bar{a}} \sum_{f \in \mathcal{B}} \max\bigl(0,\; a_f - 1.5\,\bar{a}\bigr).
\end{equation}

\noindent\textbf{Smoothness Loss.} The smoothness loss keeps the jerk within the range of real motion from both sides. Let $\mathcal{R} = \{N-2, \dots, N+K-2\}$ denote the frames whose jerk involves the transition. The loss consists of three terms,
\begin{equation}
\begin{aligned}
  \mathcal{L}_{\mathrm{up}} &= \frac{1}{\bar{r}} \max\Bigl(0,\; \frac{1}{|\mathcal{R}|} \sum_{f \in \mathcal{R}} r_f - 1.4\,\bar{r}\Bigr), \\
  \mathcal{L}_{\mathrm{low}} &= \frac{1}{\bar{r}\,|\mathcal{R}|} \sum_{f \in \mathcal{R}} \max\bigl(0,\; 0.9\,\bar{r} - r_f\bigr), \\
  \mathcal{L}_{\mathrm{lvl}} &= \frac{1}{\bar{c}\,|\mathcal{R}'|} \sum_{f \in \mathcal{R}'} \bigl|\, \mathbb{E}_{b}\,[c_f] - \bar{c} \,\bigr|,
\end{aligned}
\end{equation}
and $\mathcal{L}_{\mathrm{smooth}} = 0.1\,\mathcal{L}_{\mathrm{up}} + 3\,\mathcal{L}_{\mathrm{low}} + \mathcal{L}_{\mathrm{lvl}}$. The first term penalizes a transition whose mean jerk is higher than that of real motion, and the second penalizes frames whose jerk is lower, which prevents a frozen or floating transition. In the third term, $c_f$ is the per-frame jerk statistic of FlowMDM~\cite{barquero2024seamless}, \ie, the largest L1 norm of the jerk over all joints, $\mathbb{E}_{b}$ is the mean over the batch, $\bar{c}$ is the mean of $c_f$ over real motion, and $\mathcal{R}' = \{N-3, \dots, N+K-1\}$. It keeps the average jerk level of every frame close to that of real motion.

\noindent\textbf{Skating Loss.} The skating loss penalizes the horizontal speed of a toe while it is near the floor. For toe $k$, let $v_{f,k}$ be the length of its horizontal displacement from frame $f-1$ to frame $f$. Then
\begin{equation}
  \mathcal{L}_{\mathrm{skate}} = \frac{1}{2\,(K+1)} \sum_{f=N}^{N+K} \sum_{k} v_{f,k}\; \sigma\bigl(2\,(h_0 + 3 - y_{f,k})\bigr),
\end{equation}
where $\sigma$ is the sigmoid function, so that the weight is close to 1 for a toe lower than 3\,cm above the floor and close to 0 for a lifted toe.

\noindent\textbf{Plan Loss.} The plan loss keeps a toe still and on the floor on the frames where the contact plan $\mathbf{p}$ puts it down,
\begin{equation}
  \mathcal{L}_{\mathrm{plan}} = \frac{1}{\sum_{f,k} p_{f,k}} \sum_{f \in \mathcal{T}} \sum_{k} p_{f,k} \bigl( v_{f,k} + \max(0,\; y_{f,k} - h_0 - 3) \bigr).
\end{equation}

\noindent\textbf{Penetration Loss.} The penetration loss penalizes the depth of the two feet and the two toes below the floor, with a margin of 1\,cm,
\begin{equation}
  \mathcal{L}_{\mathrm{pen}} = \frac{1}{4K} \sum_{f \in \mathcal{T}} \sum_{k \in \mathcal{F}} \max\bigl(0,\; h_0 - 1 - y_{f,k}\bigr),
\end{equation}
where $\mathcal{F}$ contains the foot and toe joints of both legs.

\noindent\textbf{Hip Loss.} The hip loss bounds the hip height from both sides. Let $\ell_f$ be the linear interpolation of the hip height between the last prefix frame and the first suffix frame, and let $o = \frac{1}{K} \sum_{f \in \mathcal{T}} (y_{f,\mathrm{hip}} - \ell_f)$ be the mean offset of the generated hip from this line. Then
\begin{equation}
  \mathcal{L}_{\mathrm{hip}} = \frac{\max(0,\; o - o^{+})}{o^{+}} + \frac{\max(0,\; -o - o^{-})}{o^{-}},
\end{equation}
where $o^{+}$ and $-o^{-}$ are the 90th and the 10th percentile of the same offset in real motion. The first term prevents the character from floating, and the second prevents it from sinking into the floor.

\noindent\textbf{Naturalness Loss.} The naturalness loss keeps the output on the motion manifold learned by the pre-trained backbone. We add noise to the output $\hat{\mathbf{x}}_0$ at a diffusion step $t'$ randomly chosen from $\{200, 400\}$, and denoise it with a frozen copy $D_{\bar{\theta}}$ of the pre-trained backbone, without the adapters and the blending schedules,
\begin{equation}
  \mathcal{L}_{\mathrm{nat}} = \frac{1}{K d} \sum_{f \in \mathcal{T}} \bigl\| \mathrm{sg}\bigl[D_{\bar{\theta}}(\mathbf{x}_{t'}, \mathbf{c}, \mathbf{m})\bigr]_f - \hat{\mathbf{x}}_{0,f} \bigr\|^2,
\end{equation}
where $\mathrm{sg}[\cdot]$ is the stop-gradient operator and $d = 135$ is the feature dimension.

\noindent\textbf{Total Loss.} The total loss is the weighted sum of all terms,
\begin{equation}
\begin{aligned}
  \mathcal{L} = {} & 10\,\mathcal{L}_{\mathrm{reach}} + \mathcal{L}_{\mathrm{seam}} + \mathcal{L}_{\mathrm{smooth}} + 0.1\,\mathcal{L}_{\mathrm{skate}} + 0.3\,\mathcal{L}_{\mathrm{plan}} \\
  & + 3\,\mathcal{L}_{\mathrm{pen}} + 3\,\mathcal{L}_{\mathrm{hip}} + \mathcal{L}_{\mathrm{nat}}.
\end{aligned}
\end{equation}
All reference levels ($\bar{a}$, $\bar{r}$, $\bar{c}$, $h_0$, $o^{+}$, $o^{-}$ and $\boldsymbol{\sigma}_j$) are measured on the training set only.

\section{Baseline Results without Post-Processing}
\label{sec:suppl_raw}

\begin{table}[t]
\centering
\setlength{\tabcolsep}{3.2pt}
\caption{Effect of the post-processing (PP) on DoubleTake, FlowMDM and ActionPlan on LaFAN1 with $K = 30$. The rows with PP are those reported in the main paper. Air and Pen.\ are not changed by PP and are omitted.}
\label{tab:raw_pp}
\resizebox{\linewidth}{!}{%
\begin{tabular}{lccccc}
\toprule
Method & Disc.\,(\%)$\downarrow$ & Bnd$^{\mathrm{pre}}\rightarrow$ & Bnd$^{\mathrm{suf}}\rightarrow$ & Jerk$\rightarrow$ & Skate$\downarrow$ \\
\midrule
Real motion & 15.7 & 1.43 & 1.43 & 1.38 & 0.62 \\
\midrule
DoubleTake~\cite{shafir2024human} & 100.0 & 17.74 & 152.39 & 23.47 & 32.63 \\
DoubleTake~\cite{shafir2024human} + PP & 100.0 & 19.99 & 25.01 & 6.79 & 7.34 \\
\midrule
FlowMDM~\cite{barquero2024seamless} & 100.0 & 14.62 & 156.21 & 24.53 & 31.48 \\
FlowMDM~\cite{barquero2024seamless} + PP & 100.0 & 16.95 & 19.16 & 6.61 & 6.67 \\
\midrule
ActionPlan~\cite{nazarenus2026actionplan} & 100.0 & 22.01 & 182.93 & 33.24 & 55.20 \\
ActionPlan~\cite{nazarenus2026actionplan} + PP & 100.0 & 22.82 & 40.51 & 14.76 & 13.48 \\
\bottomrule
\end{tabular}}
\end{table}

DoubleTake, FlowMDM and ActionPlan generate the root as velocities, so the generated suffix drifts away from the given one, and replacing it by the given suffix leaves a large jump at the suffix boundary. In the main paper, we therefore report these three methods with a post-processing that rigidly aligns the generated suffix to the given one and blends the offset into the transition. Table~\ref{tab:raw_pp} compares the results with and without it. The post-processing lowers Bnd$^{\mathrm{suf}}$ by 4.5 to 8.2 times (\eg, 156.21 vs.\ 19.16 for FlowMDM), and also lowers Jerk and Skate by 2.3 to 4.7 times. Bnd$^{\mathrm{pre}}$ rises slightly (\eg, 14.62 vs.\ 16.95 for FlowMDM), since the blended offset also changes the first transition frames. Therefore, the numbers in the main paper are the more favorable ones for these methods, and they still jump at the boundaries in every pair (Disc of 100\%).

\section{Results for a Longer Transition}
\label{sec:suppl_k}
Table~\ref{tab:lafan_k45} and Table~\ref{tab:hml3d_k45} report the comparison with $K = 45$ (1.5\,s) frames on LaFAN1 and HumanML3D, with the same test pairs, baselines and metrics as in the main paper. HumanBridge again has the lowest discontinuity rate (0.1\% on LaFAN1 and 4.7\% on HumanML3D), together with the lowest foot skating, floor penetration, PJ and AUJ. Inertialization remains the strongest baseline at the boundaries, but its feet slide 3 to 4 times more than ours (2.46 vs.\ 0.66 on LaFAN1 and 1.28 vs.\ 0.43 on HumanML3D). Moreover, our jerk stays close to that of real motion (1.61 vs.\ 1.38 on LaFAN1), so the longer transition does not make our results stiffer.

\begin{table*}[t]
\centering
\scriptsize
\setlength{\tabcolsep}{3.2pt}
\caption{Quantitative comparison on LaFAN1 with $K = 45$, over 2219 test pairs. $\downarrow$: lower is better; $\rightarrow$: closer to real motion is better. Bnd$^{\mathrm{pre}}$, Bnd$^{\mathrm{suf}}$ and Jerk are given as multiples of the median of real motion, and Skate is in cm per frame. The best generated result is shaded in red and the second best in orange; for $\rightarrow$ columns, the closest to real motion.}
\label{tab:lafan_k45}
\resizebox{\linewidth}{!}{%
\begin{tabular}{lcccccccccc}
\toprule
Method & Disc.\,(\%)$\downarrow$ & Bnd$^{\mathrm{pre}}\rightarrow$ & Bnd$^{\mathrm{suf}}\rightarrow$ & Jerk$\rightarrow$ & Skate$\downarrow$ & SkateR\,(\%)$\downarrow$ & Air\,(\%)$\rightarrow$ & Pen.\,(\%)$\downarrow$ & PJ$\downarrow$ & AUJ$\downarrow$ \\
\midrule
Real motion & 16.4 & 1.43 & 1.44 & 1.38 & 0.62 & 17.9 & 20.1 & 0.00 & 0.031 & 0.010 \\
\midrule
Inertialization~\cite{bollo2018inertialization} & \second{0.8} & 0.52 & \first{0.87} & 0.25 & 2.46 & 31.0 & 56.2 & 11.47 & \second{0.045} & 0.313 \\
Inertialization~\cite{bollo2018inertialization} + foot lock & 19.0 & 0.52 & 2.15 & 0.68 & 1.81 & 14.2 & 64.8 & 7.22 & 0.159 & 0.557 \\
TST~\cite{qin2022motion} & 65.1 & \second{2.08} & 5.69 & 1.85 & 2.53 & 46.1 & \first{21.5} & 11.83 & 0.119 & \second{0.255} \\
\midrule
DoubleTake~\cite{shafir2024human} & 100.0 & 19.71 & 22.98 & 4.70 & 6.50 & 40.0 & 61.1 & 19.36 & 0.871 & 1.824 \\
FlowMDM~\cite{barquero2024seamless} & 100.0 & 17.36 & 19.94 & 5.02 & 6.50 & 41.2 & 62.2 & 17.56 & 0.698 & 1.622 \\
CondMDI~\cite{cohan2024flexible} & 91.1 & 3.41 & 4.79 & 3.20 & 2.23 & 43.4 & \second{21.6} & 4.40 & 0.124 & 0.364 \\
Kimodo~\cite{rempe2026kimodo} & 46.9 & 2.60 & 2.59 & 2.03 & 1.64 & \first{7.6} & 78.5 & 1.09 & 0.207 & 0.645 \\
MotionBricks~\cite{wang2026motionbricks} & 82.1 & 3.70 & 3.44 & \second{1.03} & \second{1.51} & 20.4 & 51.1 & 3.17 & 0.164 & 0.378 \\
Sched.\ Inpainting~\cite{agrawal2026interactive} & 95.9 & 6.20 & 4.53 & 3.46 & 2.36 & 45.1 & \second{21.6} & 5.47 & 0.115 & 0.350 \\
ActionPlan~\cite{nazarenus2026actionplan} & 100.0 & 26.73 & 38.35 & 9.68 & 11.92 & 27.6 & 84.0 & \second{1.08} & 1.377 & 3.161 \\
\midrule
\textbf{HumanBridge (ours)} & \first{0.1} & \first{0.80} & \second{0.83} & \first{1.61} & \first{0.66} & \second{13.4} & 18.5 & \first{0.61} & \first{0.040} & \first{0.027} \\
\bottomrule
\end{tabular}
}
\end{table*}

\begin{table*}[t]
\centering
\scriptsize
\setlength{\tabcolsep}{3.2pt}
\caption{Quantitative comparison on HumanML3D with $K = 45$, over 6650 test pairs. $\downarrow$: lower is better; $\rightarrow$: closer to real motion is better. The best generated result is shaded in red and the second best in orange; for $\rightarrow$ columns, the closest to real motion.}
\label{tab:hml3d_k45}
\resizebox{\linewidth}{!}{%
\begin{tabular}{lcccccccccc}
\toprule
Method & Disc.\,(\%)$\downarrow$ & Bnd$^{\mathrm{pre}}\rightarrow$ & Bnd$^{\mathrm{suf}}\rightarrow$ & Jerk$\rightarrow$ & Skate$\downarrow$ & SkateR\,(\%)$\downarrow$ & Air\,(\%)$\rightarrow$ & Pen.\,(\%)$\downarrow$ & PJ$\downarrow$ & AUJ$\downarrow$ \\
\midrule
Real motion & 19.9 & 1.62 & 1.73 & 1.78 & 0.37 & 7.8 & 25.7 & 0.10 & 0.013 & 0.004 \\
Real motion via HumanML3D~\cite{guo2022generating} features & 63.4 & 4.38 & 4.65 & 2.04 & 0.41 & 8.0 & 26.4 & 0.13 & 0.036 & 0.074 \\
\midrule
Inertialization~\cite{bollo2018inertialization} & \second{6.7} & 0.59 & \first{1.04} & 0.60 & 1.28 & 24.3 & 34.1 & 9.66 & \second{0.019} & \second{0.138} \\
Inertialization~\cite{bollo2018inertialization} + foot lock & 39.4 & \second{0.60} & 3.66 & \second{1.66} & \second{0.88} & 13.0 & 42.3 & 7.92 & 0.150 & 0.458 \\
\midrule
DoubleTake~\cite{shafir2024human} & 100.0 & 39.31 & 53.74 & 11.38 & 5.57 & 15.3 & 79.1 & 7.23 & 0.759 & 1.649 \\
FlowMDM~\cite{barquero2024seamless} & 100.0 & 31.04 & 42.70 & 11.16 & 4.96 & 15.1 & 82.3 & 5.32 & 0.543 & 1.319 \\
CondMDI~\cite{cohan2024flexible} & 100.0 & 11.65 & 14.65 & 6.24 & 0.91 & 10.2 & \first{22.1} & 2.01 & 0.241 & 0.625 \\
Kimodo~\cite{rempe2026kimodo} & 96.9 & 5.05 & 4.76 & 3.95 & 1.03 & \first{4.4} & 82.5 & \second{0.29} & 0.089 & 0.426 \\
MotionBricks~\cite{wang2026motionbricks} & 99.8 & 6.28 & 5.82 & \first{1.76} & 0.93 & 11.8 & \second{30.4} & 2.91 & 0.077 & 0.178 \\
ActionPlan~\cite{nazarenus2026actionplan} & 100.0 & 44.45 & 62.89 & 20.48 & 4.82 & 17.4 & 71.5 & 1.63 & 0.943 & 2.293 \\
\midrule
\textbf{HumanBridge (ours)} & \first{4.7} & \first{1.01} & \second{1.03} & 2.29 & \first{0.43} & \second{8.7} & 16.8 & \first{0.13} & \first{0.018} & \first{0.017} \\
\bottomrule
\end{tabular}
}
\end{table*}

\section{Accuracy of the Contact Planner}
\label{sec:suppl_plan}
Since no ground-truth transition exists between two different motions, we evaluate the contact planner on the 2219 LaFAN1 test windows played through without a transition, where the recorded motion between the two ends serves as the label. A toe is in contact when it is less than 5\,cm above the floor and moves slower than 1\,cm per frame. With $K = 30$, the planned contacts agree with the recorded ones in 92.3\% of the toe frames (F1 of 92.1\%), whereas simply keeping the contact state of the last prefix frame only reaches 62.1\% (F1 of 61.1\%). On the bridging pairs, the contacts of our generated transitions agree with the plan in 89.0\% of the toe frames (F1 of 88.1\%), so the generated transitions follow the plan closely.

\section{Additional Qualitative Results}
\label{sec:suppl_qual}
\begin{figure}
    \centering
    \includegraphics[width=0.98\linewidth]{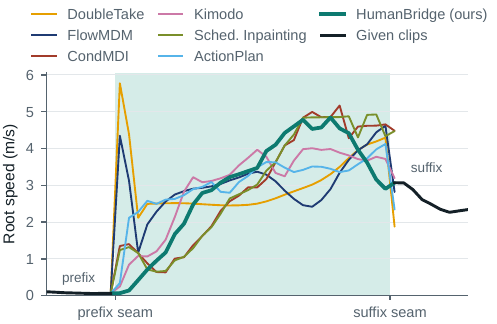}
    \caption{Root speed across the two boundaries on one LaFAN1 pair, where the character starts from standing still and ends in a run. The baselines change their speed abruptly at the boundaries, while HumanBridge speeds up smoothly inside the transition and joins both clips at their own speed.}
    \label{fig:seam_speed}
\end{figure}

\noindent\textbf{Root Speed at the Boundaries.} Figure~\ref{fig:seam_speed} shows the root speed of one pair that starts from standing still and ends in a run. DoubleTake and FlowMDM jump to 5.8 and 4.3\,m/s in the first transition frame, and ActionPlan, CondMDI and Scheduled Inpainting also change their speed within one or two frames. In contrast, HumanBridge starts as slowly as real motion and enters the suffix clip close to its speed (2.9 vs.\ 3.1\,m/s), since the blending schedules keep the transition close to each clip near its boundary.

\begin{figure*}[ht]
    \centering
    \includegraphics[width=0.98\linewidth]{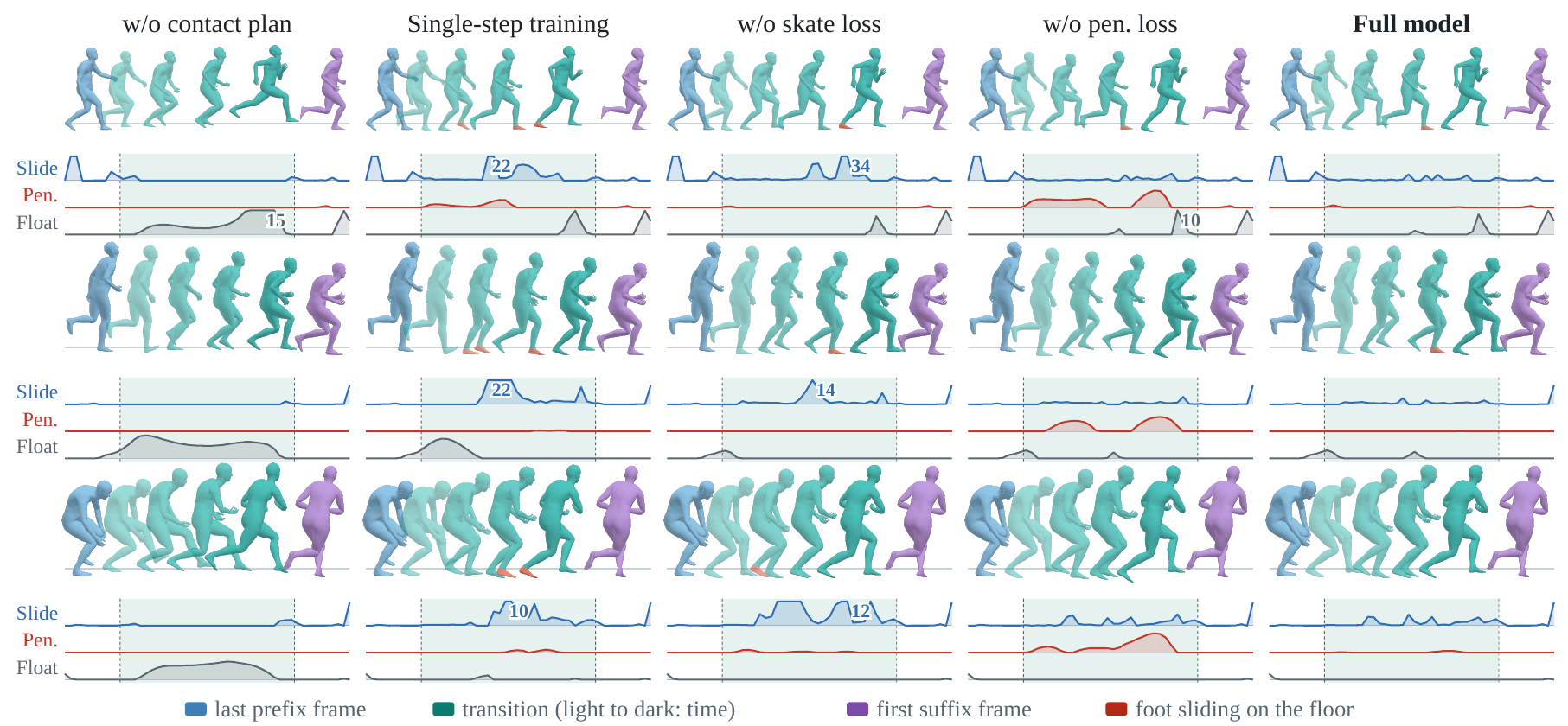}
    \caption{Qualitative ablation on three test pairs, shown as in the qualitative comparison of the main paper. The curves below show the foot sliding speed (0 to 6\,cm per frame), the floor penetration (0 to 8\,cm) and the floating height of both feet above the floor (0 to 10\,cm) per frame. Removing the contact plan lets both feet float, single-step training and removing the skating loss bring back sliding, and removing the penetration loss lets the feet sink into the floor, while the full model keeps all three low.}
    \label{fig:ablation_qual}
\end{figure*}

\noindent\textbf{Qualitative Ablation.} Figure~\ref{fig:ablation_qual} shows four ablation variants of the main paper on three LaFAN1 test pairs. Without the contact plan, both feet leave the floor for 24 to 28 consecutive frames, with the lower toe 8 to 9\,cm above the floor on average, whereas the full model keeps a foot on the floor in 24 to 30 of the 30 frames. Single-step training and removing the skating loss bring the sliding back, with a mean sliding speed of 1.20 to 3.49\,cm per frame (vs.\ 0.33 to 0.80 for the full model). Removing the penetration loss lets the toes sink 4.8 to 6.4\,cm into the floor (vs.\ at most 0.7\,cm for the full model). The variants without the blending schedules are compared quantitatively in Table~3 of the main paper.


\section{Limitations}
\label{sec:suppl_limit}
Our method has several limitations. First, on HumanML3D our transitions lift the feet less often than real motion (an Air of 17.1\% vs.\ 25.4\% with $K = 30$), so the contact plan there is more conservative than real motion. Second, the contact planner and the losses assume a flat floor and only consider the contacts of the two toes. Transitions on uneven terrain, or between motions in which other body parts touch the ground, are therefore not explicitly handled. Finally, the contact planner and the losses are tuned on motion capture data, and we leave the extension to more general contacts and to other datasets as future work.

\end{document}